\documentclass[lettersize,journal]{IEEEtran}
\usepackage{amsmath,amsfonts}
\usepackage{algorithmic}
\usepackage{algorithm}
\usepackage{array}
\usepackage{textcomp}
\usepackage{stfloats}
\usepackage{url}
\usepackage{verbatim}
\usepackage{graphicx}
\usepackage{cite}
\usepackage{multirow}
\usepackage{xcolor}
\usepackage{hyperref}
\usepackage{subcaption}
\usepackage{pifont}

\newcommand{\method}{METER}
\newcommand{\methodd}{METER }

\begin{document}

\title{\textbf{METER: a mobile vision transformer architecture for monocular depth estimation}}

\author{Lorenzo Papa~\IEEEmembership{Student Member IEEE}, Paolo Russo and Irene Amerini~\IEEEmembership{Member IEEE}
\thanks{The authors are with the  Department of Computer, Control and Management Engineering, Sapienza University of Rome, Italy (e-mail: [papa, paolo.russo, amerini]@diag.uniroma1.it).}
\thanks{Manuscript received Month Day, 2022; revised Month Day, 2022.}
}

\markboth{Journal of \LaTeX\ Class Files,~Vol.~14, No.~8, August~2021}%
{Shell \MakeLowercase{\textit{et al.}}: A Sample Article Using IEEEtran.cls for IEEE Journals}

\IEEEpubid{\begin{minipage}{\textwidth}\ \\[12pt]
Copyright~\copyright~20xx IEEE. Personal use of this material is permitted. However, permission to use this material for any other purposes must be obtained from the IEEE by sending an email to pubs-permissions@ieee.org.
\end{minipage}}


\maketitle
    
\begin{abstract}
Depth estimation is a fundamental knowledge for autonomous systems that need to assess their own state and perceive the surrounding environment. 
Deep learning algorithms for depth estimation have gained significant interest in recent years, owing to the potential benefits of this methodology in overcoming the limitations of active depth sensing systems.
Moreover, due to the low cost and size of monocular cameras, researchers have focused their attention on monocular depth estimation (MDE), which consists in estimating a dense depth map from a single RGB video frame.
State of the art MDE models typically rely on vision transformers (ViT) architectures that are highly deep and complex, making them unsuitable for fast inference on devices with hardware constraints. 

Purposely, in this paper, we address the problem of exploiting ViT in MDE on embedded devices. 
Those systems are usually characterized by limited memory capabilities and low-power CPU/GPU.
We propose \method, a novel lightweight vision transformer architecture capable of achieving state of the art estimations and low latency inference performances on the considered embedded hardwares: NVIDIA Jetson TX1 and NVIDIA Jetson Nano.
We provide a solution consisting of three alternative configurations of \method, a novel loss function to balance pixel estimation and reconstruction of image details, and a new data augmentation strategy to improve the overall final predictions.
The proposed method outperforms previous lightweight works over the two benchmark datasets: the indoor NYU Depth v2 and the outdoor KITTI.
\end{abstract}


\section{Introduction}
\label{sec:introduction}
Acquiring accurate depth information from a scene is a fundamental and important challenge in computer vision, as it provides essential knowledge in a variety of vision applications, such as augmented reality, salient object detection, visual SLAM, video understanding, and robotics~\cite{survey1, survey2, trans_depth_add_3}. 
Depth data is usually captured with active depth sensors as LiDARs, depth cameras, and other specialised sensors capable of perceiving such information by perturbing the surrounding environment, e.g. through time-of-flight or structured light technologies.
These sensors have several disadvantages, including unfilled depth maps and restricted depth ranges, as well as being difficult to integrate into low-power embedded devices. 
In addition, we also need to consider the power consumption in the case of hardwares with low-resource constraints.

On the contrary, passive depth sensing systems based on deep learning (DL) could potentially overcome all the active depth sensor limitations.
Moreover, in some settings such as indoor or hostile environments, where the use of small robots and drones could introduce additional constraints, the presence of a single RGB camera offers an effective and low-cost alternative to such traditional setups. 
The monocular depth estimation (MDE) task consists in the prediction of a dense depth map from a video frame with the use of DL algorithms, where the estimation is computed for each pixel.

Recent MDE models aim at enabling depth perception using single RGB images on deep vision transformer (ViT) architectures~\cite{dpthybrid, adabins, binsformer}, which are generally unsuitable for fast inference on low-power hardwares. 
Instead, well-established convolutional neural networks (CNN) architectures~\cite{speed, fastdepth} have been successfully exploited on embedded devices with the goal of achieving accurate and low latency inferences.
However, ViT architectures demonstrate the advantage of a global processing by obtaining significant performance improvements over fully-CNNs.
In order to balance computational complexity and hardware constraints, we propose to integrate the two architectures by fusing transformers blocks and convolutional operations, as successfully exploited in  classification and object detection~\cite{mobvit, mobformer} tasks.

\IEEEpubidadjcol
This paper presents \method, a \textit{MobilE vision TransformER} architecture for MDE that achieves state of the art results with respect to previous lightweight models over two benchmark datasets, i.e. NYU Depth v2~\cite{nyu} and  KITTI~\cite{kitti}. \methodd inference speed will be evaluated on two embedded hardwares, the 4GB NVIDIA Jetson TX1 and the 4GB NVIDIA Jetson Nano.
To improve the overall estimation performances, we focus on three fundamental components: a specific loss function, a novel data augmentation policy and a custom transformer architecture.
The loss function is composed of four independent terms
(quantitative and similarity measurements)
to balance the architecture reconstruction capabilities while highlighting the image high-frequency details.
Moreover, the data augmentation strategy employs a simultaneous random shift over both the input image and the dense ground truth depth map to increase model resilience to tiny changes of illumination and depth values.  

\begin{figure*}[t]
    \centering
    \includegraphics[width=\linewidth]{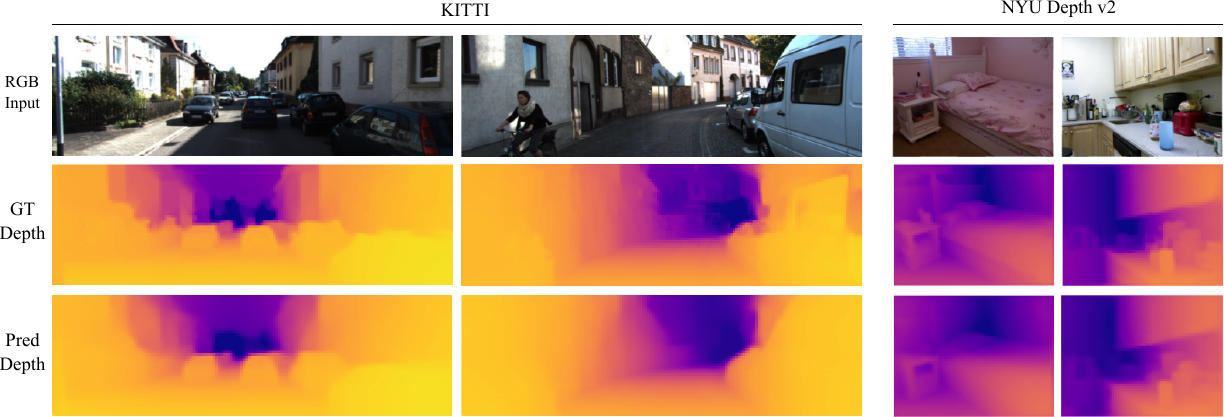}
    \caption{\methodd depth map predictions (third-row) over the KITTI and NYU Depth v2 datasets. 
    GT depth maps are resized to match \methodd output resolution.
    The depth maps are converted in RGB format with a perceptually uniform colormap (Plasma-reversed) extracted from the ground truth (second-row), for a better view.}
    \label{fig:meter_examples}
\end{figure*}

The proposed network exploits a hybrid encoder-decoder structure characterized by a ViT encoder, which was inspired by~\cite{mobvit} due to its fast inference performances.  
We focus on the transformer structure in order to identify and to improve the blocks with the highest computational cost while optimizing the model to extract robust features.
In addition, we designed a novel lightweight CNN decoder to limit the amount of operations while improving the reconstruction process. 
Furthermore, we propose three different \methodd configurations; for each variant, we reduce the number of trainable parameters at the expense of a slight increase of the final estimation error. 
Figure~\ref{fig:meter_examples} shows several \methodd depth estimations for both indoor and outdoor environments.

Moreover, to the best of our knowledge, \methodd is the first model for the MDE task that integrates the advantage of ViT architectures in such lightweight DL structures under low-resource hardware constraints.
The main contributions of the paper are summarized as follows:
\begin{itemize}
    \item We propose a novel lightweight ViT architecture for monocular depth estimation able to infer at high frequency on low-resource (4GB) embedded devices.
    \item We introduce a novel data augmentation method and loss function to boost the model estimation performances.
    \item We show the effectiveness and robustness of \methodd with respect to related state of the art MDE methods over two benchmark datasets, i.e. NYU Depth v2~\cite{nyu} and KITTI~\cite{kitti}.
    \item We validate the models through quantitative and qualitative experiments, data augmentation strategies and a loss function components, highlighting their effectiveness.
\end{itemize}

This paper is organized as follows: Section~\ref{sec:related_works} reviews some previous works related to the topics of interest. Section~\ref{sec:proposed_method} describes the proposed method and the overall architecture in detail.
Experiments and hyper-parameters are discussed in Section~\ref{sec:experimental_setup}, while Section~\ref{sec:results}  reports the results and a quantitative analysis of \methodd with respect to other significant works. Some final considerations and future applications are provided in Section~\ref{sec:conclusion}.

\section{Related Works}
\label{sec:related_works}
In this section, we report state of the art related works on monocular depth estimation, grouped as follows: fully CNN-based methods are covered in Section~\ref{subsec:rw_cnn}, ViT-based approaches in Section~\ref{subsec:rw_vit} and lightweight (CNN) MDE methods in Section~\ref{subsec:rw_fast_cnn}. 

\subsection{CNN-based MDE methods}
\label{subsec:rw_cnn}
Fully convolutional neural networks based on encoder-decoder structures are commonly used for dense prediction tasks such as depth estimation and semantic segmentation. 
In the seminal work of Eigen et al.~\cite{depth_first} it is presented a CNN model to handle the MDE task by employing two stacked deep networks to extract both global and local informations.
Cao et al. present~\cite{trans_depth_add_1} and~\cite{trans_depth_add_2} two works based on deep residual networks to solve the MDE defined as a classification task, respectively, over absolute and relative depth maps.
Alhashim et al.~\cite{densedepth} propose DenseDepth, a network which exploits transfer learning to produce high-resolution depth maps. The architecture is composed of a standard encoder-decoder with a pre-trained DenseNet-169~\cite{densenet} as backbone and a specifically designed decoder. 
Gur et al.~\cite{deeplab} present a variant of the DeepLabV3+~\cite{deeplabv3} model where the encoder is composed of a ResNet~\cite{resnet} and of an \textit{atrous} spatial pyramidal pooling while introducing a Point Spread Function convolutional layer to learn depth informations from defocus cues.
Recently, Song et al.~\cite{lapdepth} propose LapDepth, a Laplacian pyramid-based architecture composed of a pretrained ResNet-101 encoder and a Laplacian pyramid decoder that combined the reconstructed coarse and fine scales to predict the final depth map.  

However, those methods, which often rely on deep pre-trained encoders and high-resolution images as input, are unsuitable for inferring on low-resource hardwares. 
In contrast, we propose a lightweight architecture that takes advantage of transformers blocks to balance global feature extraction capabilities and the overall computational complexity of convolutional operations.

\subsection{ViT-based MDE methods}
\label{subsec:rw_vit}
Vision Transformers~\cite{vit_classification} gain popularity for their accuracy capabilities thanks to the attention mechanism~\cite{attention_vit} that simultaneously extract information from the input pixels and their inter-relation, outperforming the translation-invariant property of convolution. 
In dense prediction tasks, ViT architectures share the same encoder-decoder structure that has significantly contributed to face many CNN vision-related problems.
Bhat et al.~\cite{adabins} have been the first to handle the MDE task with ViT architectures by proposing Adabins: it uses a minimized version of a vision transformer structure to adaptively calculate bins width.
Ranftl et al.~\cite{dpthybrid} investigate the application of ViT proposing DPT, a model composed of a transformer-CNN encoder and a fully-convolutional decoder.
The authors show that ViT encoders provide finer-grade predictions with respect to standard CNNs, especially when instantiated with a large amount of training data.
Yun et al.~\cite{ref_rev2} improves $360^{\circ}$ monocular depth estimation methods with a joint supervised and self-supervised learning strategies taking advantage of non-local DPT.
Recently, Li et al.~\cite{trans_depth_add_4} design MonoIndoor++, a framework that takes in account the main challenges of indoor scenarios.
Kim et al.~\cite{glpdepth} propose GLPDepth, a global-local transformer network to extract meaningful features at different scales and a Selective Feature Fusion CNN block for the decoder.
The authors also integrate a revisited version of CutDepth data augmentation method~\cite{cutdepth} which is able to improve the training process on the NYU Depth v2 dataset without needing additional data.
Li et al. propose DepthFormer~\cite{binsformer} and BinsFormer~\cite{depthformer}, where the first one is composed of a fully-transformer encoder and a convolutional decoder interleaved by an interaction module to enhance transformer encoded and CNN decoded features.
Differently, in BinsFormer the idea of the authors is to use a multi-scale transformer decoder to generate adaptive bins and to recover spatial geometry information from the encoded features. 

Instead of following the recent trend of high-capacity models, we propose a novel lightweight ViT architecture that is able to achieve accurate, low latency depth estimations on embedded devices.

\subsection{Lightweight MDE methods}
\label{subsec:rw_fast_cnn}
The models reported so far are not suitable for embedded devices due to their size and complexity.  
For this reason, developing lightweight architectures could be a solution to perform inference on constrained hardwares as shown in~\cite{lite_arch_1, lite_arch_2}. 
To provide a clearer overview of those approaches we also provide the frames per second (fps) published in the original papers that focus on inference frequency, remarking that they are not comparable due to the different tested hardwares. 
Poggi et al.~\cite{PyD-Net} propose PyD-Net, a pyramidal network to infer on CPU devices. 
The authors use the pyramidal structure to extract features from the input image at different levels, which are afterwards upsampled and merged to refine the output estimation. Such model achieves less than $1$ fps on an ARM CPU and almost $8$ fps on an Intel i7 CPU.
Spek et al.~\cite{cream} present CReaM, a fully convolutional architecture obtained through a knowledge-transfer learning procedure. The model is able to achieve real-time frequency performances ($30$ fps) on the 8GB NVIDIA Jetson TX2 device.  
Wofk et al.~\cite{fastdepth} develop FastDepth, an encoder-decoder architecture characterized by a MobileNet \cite{mobnetv2} pre-trained network as backbone, and a custom decoder. 
Furthermore, the authors show that pruning the trained model guarantees a boost of inference frequency at the expense of a small increment of the final estimation error. FastDepth achieves $178$ fps on the 8GB NVIDIA Jetson TX2 device. 
Recently, Yucel et al.~\cite{trasfer_l} propose a small network composed by the MobileNet v2~\cite{mobnetv2} as encoder and FBNet x112~\cite{fbnet} as decoder, trained on an altered knowledge distillation process; the model achieves $37$ fps on smartphone GPU.
Papa et al.~\cite{speed} design SPEED, a separable pyramidal pooling architecture characterized by an improved version of the MobileNet v1~\cite{mobnetv1} as an encoder and a dedicated decoder. This architecture exploits the use of depthwise separable convolutions, achieving real-time frequency performances on the embedded 4GB NVIDIA Jetson TX1 and $6$ fps on the Google Dev Board Edge TPU.

As previously mentioned, all those lightweight MDE works are designed over fully-convolutional architectures. 
In contrast to previous methodologies, \methodd exploits a lightweight transformer module in three different configurations, achieving state of the art results over the standard evaluation metrics.

\begin{figure*}[ht]
    \centering
    \includegraphics[width=\linewidth]{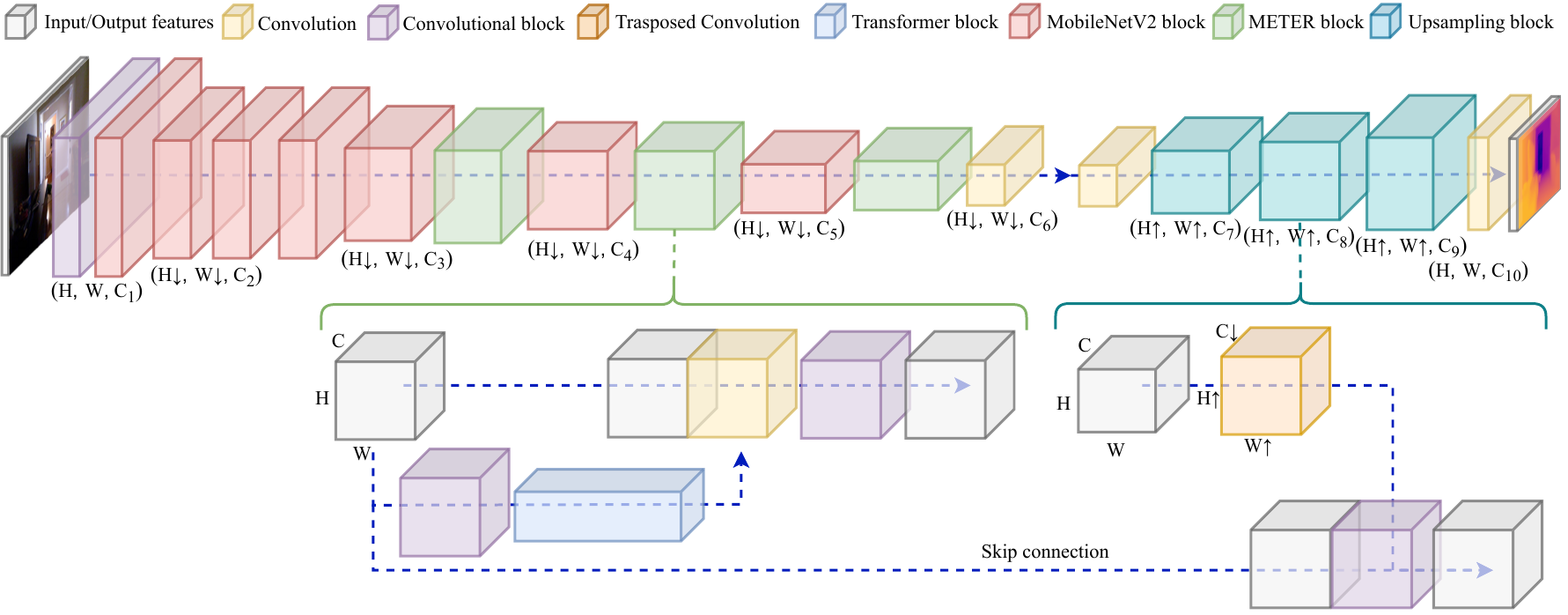}
    \caption{Overview of \methodd encoder-decoder network structure. The processing flow, i.e. the sequence of operations and the skip-connection, is represented with a blue dashed arrow. The (H, W, C) format refers to the input-output spatial dimensions, while the $\uparrow$ and $\downarrow$ refers to the feature resolution upsampling and downsampling.}
    \label{fig:meter}
\end{figure*}

\section{Proposed Method}
\label{sec:proposed_method}
This section outlines the design of METER, the proposed lightweight monocular depth estimator. 
In particular, in Section~\ref{subsec:architecture_details}, we provide a detailed architecture analysis for both encoder and decoder modules, in Section~\ref{subsec:loss_function} we describe the proposed loss function and in Section~\ref{subsec:combined_augmentation} the employed augmentation policy.

\subsection{\methodd architecture}
\label{subsec:architecture_details}
The vision transformer architecture has demonstrated outstanding performances in a variety of computer vision tasks, usually relying on deep and heavy structures. 
On the other hand, to reduce the computational cost of such models, lightweight CNN usually relies on convolutional operations with small kernels (i.e. 3x3, 1x1) or on particular techniques such as depthwise separable convolution~\cite{depth_w_conv}.
Based on those statements, we design an hybrid lightweight ViT characterized by convolutions with small kernels and as few transformers blocks as possible reducing the computational impact in the overall structure. 
Motivated by this, in the following, we present \method: a \textit{MobilE vision TrasformER} architecture characterized by a lightweight encoder-decoder model designed to infer on embedded devices.
\methodd encoder re-design computational demanding operations of~\cite{mobvit} to improve the inference performances while maintaining the feature extraction capabilities.
The high-level features extracted from the encoder are then fed into the decoder through the skip-connections to recover the image details. 
The proposed fully convolutional decoder has been structured to upsample the compact set of encoder high-level features while enhancing the reconstruction of the image details to obtain the desired output depth map (i.e. a per-pixel distance map).
A graphical overview of the architecture is reported in Figure~\ref{fig:meter} while the number of channels employed in the different \methodd configurations, \methodd S, \methodd XS, and \methodd XXS are reported in Table~\ref{tab:conv_filter}. 
The number of trainable parameters of the three proposed networks consist of $3.29M$, $1.45M$, and $0.71M$, respectively.

\begin{table}[h]
  \centering
  \caption{Number of channels ($C_i$) used in \methodd configurations.}
  \label{tab:conv_filter}
  \begin{tabular}{ c | c c c }
    Channels & \methodd S & \methodd XS & \methodd XXS \\
    \hline
    $C_1$ & 16 & 16 & 16 \\
    $C_2$ & 32 & 32 & 16 \\
    $C_3$ & 64 & 48 & 24 \\
    $C_4$ & 128 & 80 & 64 \\
    $C_5$ & 160 & 96 & 80 \\
    $C_6$ & 320 & 192 & 160 \\
    $C_7$ & 128 & 128 & 64 \\
    $C_8$ & 64 & 64 & 32 \\
    $C_9$ & 32 & 32 & 16 \\
    $C_{10}$ & 16 & 16 & 8 \\
  \end{tabular}
\end{table}

\vspace{1em}
\noindent
\textbf{\methodd encoder} exploits a modified version of MobileViT network due to its light structure demonstrated in \cite{mobvit}. 
As can be noticed in Figure~\ref{fig:meter}, \methodd presents a hybrid network composed of convolutional MobileNetV2 blocks (red) and transformers blocks (green).
The MobileViT blocks with the  highest computational cost,
i.e. the ones composed of cascaded transformers and convolution operations, have been identified and replaced with new modules (\methodd blocks).
Such modules are able to guarantee low latency inference while tuning the entire structure to minimize the final estimation error. 
Along the lines of~\cite{mobvit}, we propose three variants of the same encoder architecture with decreasing complexity and computational cost namely $S$, $XS$, and $XXS$. 

The proposed \methodd block (green in Figure \ref{fig:meter}) is composed by three feature extraction operations, two Convolutional blocks composed by a $3\times3$ convolution and a point-wise one (purple) and a second $1\times1$ convolution (yellow) interleaved by a single transformer block (blue).
Such module computes an unfold operation to apply the transformer attention on the flattened input patches while reconstructing output feature map with an opposite folding operation, as described in~\cite{mobvit}.
Moreover, in order to apply an attention mechanism to the encoded features, the input of \methodd block (gray) has been concatenated  with the output of the transformer and fed to the previous $1\times1$ convolution layer.
When compared with MobileViT architecture, characterized by four convolutions operations and a number of cascaded transformers blocks, the proposed design allows to reduce the computational cost of the overall model while producing an accurate estimation of the depth (as will be shown in Section~\ref{subsec:architecture_ablation}).

Finally, we halved the number of output encoder features (channel $C_6$) and we replaced the MobileViT SiLU non linearity function with the ReLU.
Despite the fact that SiLU activation function is differentiable at every point\footnote{Unlike the SiLU, the ReLU activation function is non-differentiable at zero.}, it does not ensure better performance, likely due to the depth-data distribution.

\vspace{1em}
\noindent
\textbf{\methodd decoder} is designed with a fully convolutional structure to enhance the estimation accuracy and the reconstruction capabilities  while keeping a limited number of operations.
As can be seen in Figure~\ref{fig:meter}, the decoder consists of a sequence of three cascaded upsampling blocks (light blue) and two convolutional layers (yellow) located at the beginning and at the end of the model. 
Each upsampling block is composed by a sequence of upsampling, skip-connection and feature extraction operations.
The upsampling operation is performed by a transposed convolutional layer (orange)  which doubles the spatial resolution of the input. Then, a Convolutional block (purple) is used for feature extraction; the skip-connection (dashed blue arrow) linking \methodd encoder-decoder modules allows to recover image details from the encoded feature maps.

\subsection{The balanced loss function}
\label{subsec:loss_function}
The standard monocular depth estimation formulation consider as loss function the per-pixel difference between the $i^{th}$ ground truth pixel $y_i$ and the predicted one $\hat{y}_i$. 
However, as reported in literature~\cite{acc_obj_loss, densedepth, loss_ssim} several modifications have been proposed to improve the convergence speed and the overall depth estimation performances. 
In particular, the addition of different loss components focuses on refinement of fine details in the scenes, like object contours.


Derived from~\cite{acc_obj_loss, loss_ssim}, we propose a \textit{balanced loss function} (BLF) to weight the reconstruction loss through the $L_{depth}(y_i, \hat{y}_i)$ and $L_{SSIM}(y_i, \hat{y}_i)$ components with the high-frequency features taken into account by the $L_{grad}(y_i, \hat{y}_i)$ and the $L_{norm}(y_i, \hat{y}_i)$ losses.
The BLF $L(y_i, \hat{y}_i)$ mathematical formulation is reported in Equation~\ref{eq:loss}, where $\lambda_1, \lambda_2, \lambda_3$ are used as scaling factors.
\begin{equation}
    \label{eq:loss}
    L(y_i, \hat{y}_i) = L_{depth} + \lambda_1 L_{grad} + \lambda_2 L_{norm} + \lambda_3 L_{SSIM}
\end{equation}


In detail, the loss $L_{depth}(y_i, \hat{y}_i)$ in Equation~\ref{eq:l_depth} is the point-wise L1 loss computed as the per-pixel absolute difference between the ground truth $y_i$ and the predicted image~$\hat{y}_i$.
\begin{equation}
    L_{depth}(y_i, \hat{y}_i) = \frac{1}{n} \sum^{n}_{i=1} |y_i - \hat{y}_i|
    \label{eq:l_depth}
\end{equation}

The $L_{grad}(y_i, \hat{y}_i)$ and the $L_{norm}(y_i, \hat{y}_i)$ losses reported respectively in Equation~\ref{eq:l_grad} and Equation~\ref{eq:l_norm}  are designed to penalize the estimation errors around the edges and on small depth details. 
The $L_{grad}(y_i, \hat{y}_i)$ loss computes the Sobel gradient function to extract the edges and objects boundaries.
\begin{equation}
    L_{grad}(y_i, \hat{y}_i) = \frac{1}{n} \sum^{n}_{i=1} (\nabla_x(|y_i - \hat{y}_i|) + \nabla_y(|y_i - \hat{y}_i|))
    \label{eq:l_grad}
\end{equation}
We report with $\nabla$ the spatial derivative of the absolute estimation error with respect to the $x$ and $y$  axes.

The $L_{norm}(y_i, \hat{y}_i)$ loss,  reported in Equation ~\ref{eq:l_norm},  calculates the cosine similarity~\cite{cos_similarity} between the ground truth and the prediction.
\begin{equation}
    L_{norm}(y_i, \hat{y}_i) = \frac{1}{n} \sum_{i=1}^n \left( 1 - \frac{\langle n_{\hat{y}_i}, n_{y_i}\rangle}{ \sqrt{\langle n_{\hat{y}_i}, n_{\hat{y}_i}\rangle} \sqrt{\langle n_{y_i}, n_{y_i}\rangle}} \right)
    \label{eq:l_norm}
\end{equation}

We identify with $\langle n_{y_i}, n_{\hat{y}_i}\rangle$ the inner product of the surface normal vectors $n_{y_i}$ and $n_{\hat{y}_i}$ computed for each depth map i.e. $n_{z}=[-\nabla_x(z),-\nabla_y(z), 1]^T$ with $z=[y_i, \hat{y}_i]$.

The last component $L_{SSIM}(y_i, \hat{y}_i)$ loss, Equation~\ref{eq:l_ssim}, is based on the mean structural similarity ($SSIM$) \cite{ssim}. Similarly to \cite{loss_ssim, densedepth} we add this function to improve the depth reconstruction and the overall final estimation.
\begin{equation}
    L_{SSIM}(y_i, \hat{y}_i) = 1 - SSIM(y_i, \hat{y}_i)
    \label{eq:l_ssim}
\end{equation}

In conclusion, the proposed BLF balances the image reconstruction $L_{depth}$, the image similarity $L_{SSIM}$, the edge reconstruction $L_{grad}$ and the edge similarity $L_{norm}$ losses.
The impact of each loss will be quantitatively evaluated in Section~\ref{subsec:loss_ablation}.


\subsection{The data augmentation policy}
\label{subsec:combined_augmentation}
Deep learning architectures and especially Vision Transformer need a large amount of input data to avoid overfitting of the given task.
Those models are typically trained on large-scale labelled datasets in a supervised learning strategy \cite{dpthybrid}. 
However, gathering annotated images is time-consuming and labour-intensive; 
as result, the data augmentation (DA) technique is a typical solution 
for expanding the dataset by creating new samples. 
In the MDE task, the use of DA techniques characterized by geometric and photometric transformations are a standard practice \cite{densedepth, adabins}.
However, not all the geometric and image transformations would be appropriate due to the introduced distortions and aberrations in the image domain, which are also reflected on the ground-truth depth maps.

With \methodd we propose a data augmentation policy based on commonly used DA operations while introducing a novel approach named \textit{shifting strategy}. 
In particular we consider as \textit{default} augmentation policy the use of the vertical flip, mirroring, random crop and channels swap of the input image as in~\cite{densedepth} to make the network invariant to specific color distributions. 
The key idea is to combine the \textit{default} augmentation policy with the \textit{shifting strategy} augmentation, based on two simultaneous transformations applied respectively to the input image and to the ground truth depth map.
The first one applies a color (\textit{C}) shift to the RGB input images, while the second one is a depth-range (\textit{D}) shift, which consists of adding a small, random positive or negative value to the depth ground truth.
The mathematical formulation of the computed transformations are following reported; we refer with $rgb_{un}$ and $rgb_{aug}$ respectively the unmodified and the augmented input for RGB images and with $d_{un}$ and $d_{aug}$ the unmodified and the augmented depth map.

The \textit{C} shift augmentation, applied on RGB images, is composed of two consecutive steps. In the first operation we apply a gamma-brightness transformation ($rgb_{gb}$), as reported in Equation~\ref{eq_gamma_da}, where $\beta$ and $\gamma$ are respectively the brightness and gamma factors that are randomly chosen into a value range experimentally defined between $[0.9, 1.1]$. 
\begin{equation}
    \label{eq_gamma_da}
    rgb_{gb} = \beta * (rgb_{un})^{\gamma} 
\end{equation}

Then, the color augmentation transformation reported in Equation~\ref{eq_color_da} is applied, where $I$ is an identity matrix of $H \times W$ resolution and $\eta$ is a scaling factor that is randomly chosen into a value range empirically set between $[0.9, 1.1]$.
\begin{equation}
    \label{eq_color_da}
    rgb_{aug} = rgb_{gb} * (I_{H \times W} * \eta)
\end{equation}

The \textit{D} shift augmentation, Equation~\ref{eq_D_aug}, is made up of a random positive or negative value summed to the ground-truth depth maps ($d_{un}$). The random value, with a range of $[-10, +10]$ centimeters for the indoor dataset and $[-10, +10]$ decimeters for the outdoor one, is uniformly applied to the whole depth map.
\begin{equation}
    \label{eq_D_aug}
    d_{aug} = d_{un} + S_{H \times W}
\end{equation}

\begin{figure}[t]
    \centering
    \includegraphics[width=\linewidth]{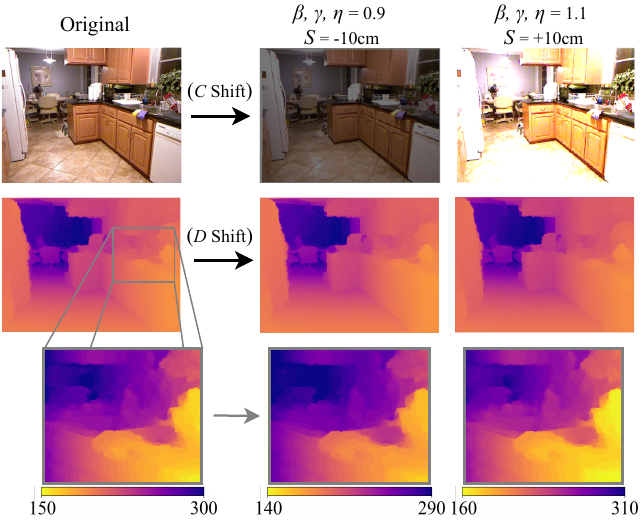}  
    \caption{Illustration of an augmented sample with the proposed \textit{shifting strategy}. The shifting factors ($\beta$, $\gamma$, $\eta$, and $S$) are set as their maximum and minimum values, i.e. $\{0.9, -10\}$ and $\{1.1, +10\}$ respectively.
    The min/max depth ranges for the regions of interest are given through the respective colored bars.}
    \label{fig:augmentation}
\end{figure}

In Figure~\ref{fig:augmentation} we report a sample frame before and after the application of the proposed strategy with the minimum and the maximum shift values.
To emphasise the impact of the \textit{D} shift, we focus on a narrow portion of the original depth map (in a distance range between $150$ and $300$ centimeters) by applying a perceptually uniform colormap and highlighting the minimum and maximum depth intervals through the associated color bars.
The reported frames show that the depth with the positive displacement ($+10$ centimeters) has a lighter colormap, while the depth with the negative displacement ($-10$ centimeters) has a darker one; this effect is emphasised by the colormap of the original distance distribution.

The introduced depth-range shift augmentation, along with the color and brightness shift and the commonly used transformations, leads to better final estimations as will be shown in Section~\ref{subsec:DA_ablation} providing also invariance to color and illumination changes.

\section{Experimental Setup}
\label{sec:experimental_setup}
This section gives a detailed description of the experimental setup, including training hyper-parameters, benchmark datasets and evaluation metrics respectively in Sections~\ref{subsec:hp},~\ref{subsec:datasets}, and~\ref{subsec:evaluation}.

\subsection{Training hyper-parameters}
\label{subsec:hp}
\methodd has been implemented using PyTorch\footnote{Code and corresponding pre-trained weights are made publicly available at the following GitHub repository: \\ \href{https://github.com/lorenzopapa5/METER}{https://github.com/lorenzopapa5/METER}} deep learning API, randomly initializing the weights of the architectures.
All the models have been trained from scratch using the AdamW optimizer \cite{adamw} with $\beta_1 = 0.9$, $\beta_2 = 0.999$, weight decay $wd = 0.01$ and an initial learning rate of $0.001$ with a decrement of $0.1$ every $20$ epochs. 
We use a batch size of $128$ for a total of $60$ epochs.   
For the \textit{balanced loss function} we empirically choose the scaling factors $\lambda_1 = 0.5$ and $\lambda_2, \lambda_3 = \{1, 10, 100\}$ depending on the unity of measure used for the predicted depth map, i.e. meters, decimeters or centimeters.
We apply a probability of $0.5$ for all the random transformations set in the data augmentation policy. 

\subsection{Benchmark datasets}
\label{subsec:datasets}
The datasets used to show the performance of \methodd are NYU Depth v2~\cite{nyu} and KITTI~\cite{kitti}, two popular MDE benchmark datasets for indoor and outdoor scenarios.

NYU Depth v2 dataset provides RGB images and corresponding depth maps in several indoor scenarios captured at a resolution of $640\times480$ pixels. The depth maps have a maximum distance of $10$ meters. The dataset contains $120K$ training samples and $654$ testing samples; we used for training the $50K$ subset as performed by previous works \cite{densedepth, adabins}. The input images have been downsampled at a resolution of $256\times192$.

KITTI dataset provides stereo RGB images and corresponding 3D laser scans in several outdoor scenarios. The RGB images are captured at a resolution of $1241\times376$ pixels. 
The depth maps have a maximum distance of $80$ meters. 
We train our network at a input resolution of $636\times192$ on Eigen et. al~\cite{depth_first} split; it is composed of almost $23K$ training and $697$ testing samples. Similarly to~\cite{lapdepth}, due to the low density depth maps, we evaluate the compared models in the cropped area where point-cloud measurement are reported.

\subsection{Performance evaluation}
\label{subsec:evaluation}
We quantitatively evaluate the performance of \methodd using common metrics \cite{depth_first} in the monocular depth estimation task: the root-mean-square error (RMSE, in meters [m]), the relative error (REL), and the accuracy value $\delta_1$, respectively reported in Equations~\ref{eq:rmse},~\ref{eq:rel}, and~\ref{eq:delta1}. We remind that $y_i$ is the ground truth depth map for the $i^{th}$ pixel while $\hat{y}_i$ is the predicted one, $n$ is the total number of pixels for each depth image, and $thr$ is a threshold commonly set to $1.25$.

\begin{equation}
    RMSE = \sqrt{\frac{1}{|n|} \sum_{i \in n} || y_i - \hat{y}_i ||^2}
    \label{eq:rmse}
\end{equation}
\begin{equation}
    REL = \frac{1}{|n|} \sum_{i \in n} \frac{|y_i - \hat{y}_i |}{y_i}
    \label{eq:rel}
\end{equation}
\begin{equation}
    \delta_1 = \frac{1}{|n|} \sum_{i \in n} \max \left(\frac{y_i}{\hat{y}_i}, \frac{\hat{y}_i}{y_i}\right) < thr 
    \label{eq:delta1}
\end{equation}

Moreover, we compare the different models through the number of multiply-accumulate (MAC) operations and trainable parameters. 
\methodd has been tested on the low-resource embedded 4GB NVIDIA Jetson TX1\footnote{\href{https://developer.nvidia.com/embedded/jetson-tx1}{https://developer.nvidia.com/embedded/jetson-tx1}} and the 4GB NVIDIA Jetson Nano\footnote{\href{https://developer.nvidia.com/embedded/jetson-nano}{https://developer.nvidia.com/embedded/jetson-nano}} that have a power consumption of $10W$ and $5W$ respectively.
Those devices are equipped with an ARM CPU and a 256-core NVIDIA Maxwell GPU\footnote{\href{https://developer.nvidia.com/maxwell-compute-architecture}{https://developer.nvidia.com/maxwell-compute-architecture}} for the TX1 and a 128-core for the Nano.
The inference speed reported in Section~\ref{sec:results} are computed as frame-per-second (fps) on a single image averaged over the entire test dataset.

\begin{figure*}[t]
    \centering
    \includegraphics[width=\linewidth]{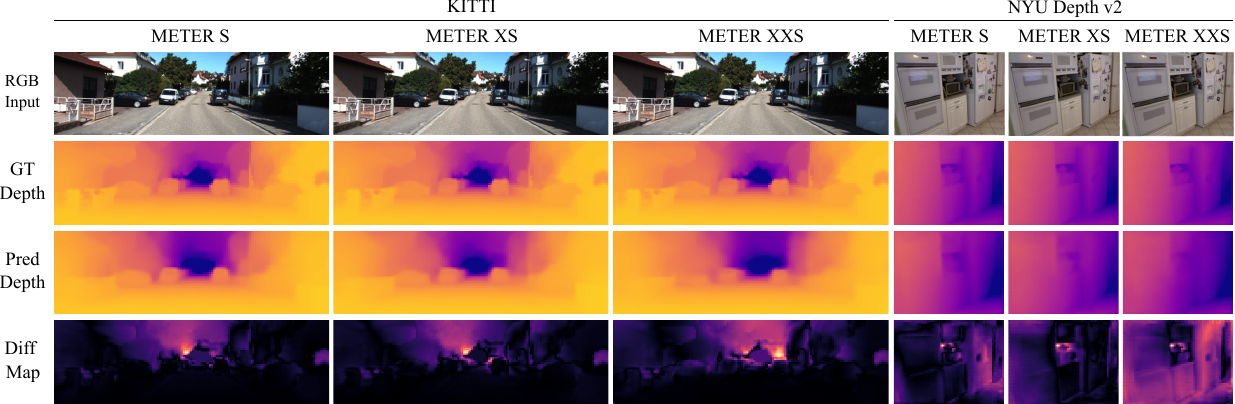}
    \caption{A graphical comparison among \methodd (S, XS, XXS) configurations. For a better visualization, we apply to depth images and difference maps uniform colormaps with the same depth range. Precisely, in the ground truth (GT) and predicted depth maps (Pred) a lower color intensity corresponds to further distances, while in the difference map (Diff $= |$GT $-$ Pred$|$) a lower color intensity corresponds to a smaller error.}
    \label{fig:meter_s_xs_xxs_comparison}
\end{figure*}

\begin{table*}[t]
  \centering
  \caption{Comparison with state of the art lightweight methods on the two benchmark datasets. The best scores are in bold and second best are underlined; the - represents a value which is not reported in the original paper.}
  \label{tab:architectures_comparison}
  \begin{tabular}{ l | c c c c | c c c c | c }
    \multirow{3}*{Models} & \multicolumn{4}{c|}{NYU} & \multicolumn{4}{c|}{KITTI} & \multicolumn{1}{c}{} \\
    & RMSE$\downarrow$ & \multirow{2}*{REL$\downarrow$} & \multirow{2}*{$\delta_1\uparrow$} & MAC & RMSE$\downarrow$ & \multirow{2}*{REL$\downarrow$} & \multirow{2}*{$\delta_1\uparrow$} & MAC & Parameters \\
     & [m]  & & & [G] & [m]  & & & [G] & [M] \\
    \hline
    CReaM \cite{cream} & 0.687 & 0.190 & 0.704 & - & - & - & - & - & - \\
    PyD-Net (50) \cite{PyD-Net} & - & - & - & - & 6.253 & 0.262 & 0.759 & - & 1.9 \\
    PyD-Net (200) \cite{PyD-Net} & - & - & - & - & 6.030 & 0.153 & 0.789 & - & 1.9 \\
    FastDepth \cite{fastdepth} & 0.579 & - & 0.772 & 3.210 & - & - & - & - & 3.9 \\
    M.Net v2 + FBNet \cite{trasfer_l} & 0.564 & - & 0.790 & - & - & - & - & - & 2.6 \\
    SPEED \cite{speed} & 0.566 & 0.158 & 0.783 & 0.552 & 5.191 & 0.181 & 0.770 & 1.403 & 2.6 \\
    \textbf{\method} S & \textbf{0.471} & \textbf{0.134} & \textbf{0.831} & 0.975 & \textbf{4.603} & \textbf{0.126} & \textbf{0.829} & 2.432 & 3.3 \\
    \textbf{\methodd XS} & \underline{0.522} & \underline{0.154} & \underline{0.793} & 0.579 & \underline{4.671} & \underline{0.128} & \underline{0.827} & 1.444 & 1.4 \\
    \textbf{\methodd XXS} & 0.580 & 0.174 & 0.744 & 0.186 & 5.157 & 0.156 & 0.782 & 0.464 & 0.7 \\
  \end{tabular}
\end{table*}

\section{Results}
\label{sec:results}
In this section, we report the results obtained with \methodd on the two evaluated datasets, NYU Depth v2 and KITTI, described in the previous Section \ref{subsec:datasets}. 
In Section~\ref{subsec:comparison_sota} \methodd is compared with lightweight, state of the art related works in terms of the metrics described in Section~\ref{subsec:evaluation}; then, we report multiple ablation studies to emphasize the individual contribution of each \methodd component. 
In particular, Section~\ref{subsec:architecture_ablation} is related to the architecture structure, while Sections~\ref{subsec:loss_ablation} and~\ref{subsec:DA_ablation} analyze respectively the effect of each element of the proposed \textit{balanced loss function} and of the \textit{shifting strategy} used for data augmentation.
Finally, in Section\ref{subsec:real_case_usage}, we provide an example of \methodd application in a real-case scenario.

\subsection{Comparison with state of the art methods}
\label{subsec:comparison_sota}
In this section, \methodd is compared with state of the art lightweight models as~\cite{speed, fastdepth, cream, PyD-Net, trasfer_l}, which are designed to infer at high speed on embedded devices while keeping a small memory footprint (lower than 3GB). 
This choice is due to the limited amount of available memory in the chosen platforms.
Usually a portion of available RAM is reserved for the operating system, thus lowering the overall amount of available space for the model allocation.
In particular \methodd and its variants allocate less than $2.1$GB of available memory, a value that does not saturate the hardware's memory and which gives the opportunity to perform other operations on the same device.
Moreover, for each compared architecture we also report the number of trainable parameters (in million [M]) and the number of Multiply-And-Accumulate (MAC) operations (in giga [G]). 

The results can be found in Table~\ref{tab:architectures_comparison}; as can be noticed, \methodd outperforms all the other methods on both the datasets.
When compared with~\cite{speed}, \methodd S achieves a boost of $17\%$, $15\%$, and $6\%$ respectively for the RMSE, REL and $\delta_1$ metrics over NYU Depth v2 dataset and of $11\%$, $30\%$ and $7\%$ over KITTI.
As before, \methodd XS achieves superior performances, with a boost of $9\%$, $10\%$ and $5\%$ over NYU Depth v2 dataset and of $10\%$, $29\%$ and $7\%$ over KITTI.
The last configuration, \methodd XXS, can still obtain good predictions compared with  state of the art models while using just  $0.7$M trainable parameters and $0.186$G MAC operations.

Moreover, in order to assess the  frequency performances of such architectures, we choose as baseline models SPEED, due to its accuracy, and FastDepth, which is one of the most popular technique.
When tested on the NVIDIA Jetson TX1, such models achieve $30.9$ fps and $18.8$ fps, while \methodd S, XS and XXS achieve respectively $16.3$ fps, $18.3$ fps and $25.8$ fps. From these results we can remark that our most accurate model shows similar fps values with respect to FastDepth with a sensible lower estimation error, while the lightweight XXS variant exhibits comparable estimation performance and fps with respect to SPEED. 

Regarding MAC operations, it is possible to see that SPEED MAC value is on par with \methodd XS, while FastDepth MAC is sensible higher than all \methodd architectures. 

Furthermore, a qualitative analysis between the proposed variants of \methodd is reported in Figure \ref{fig:meter_s_xs_xxs_comparison} over an indoor and outdoor scenarios. 
The estimated depths and their associated difference (Diff) maps, which are per-pixel differences between the ground truth depth maps (GT Depth) and the predicted (Pred Depth) ones, show how the estimation error is distributed along the frame.
Precisely, we notice an error increment fairly distributed over the frame as the model trainable parameters of the model are reduced.


\begin{table*}[!t]
  \centering
  \caption{Comparison between the MobileViT \cite{mobvit} and \methodd encoders over different activation functions (ReLU, SiLU) keeping \methodd decoder fixed. The fps are measured on the two benchmark hardwares, the NVIDIA Jetson TX1 and the NVIDIA Jetson Nano. In bold the best results for each configuration in terms of RMSE, REL and $\delta_1$.}
  \label{tab:encoder_comaprison}
  \begin{tabular}{ l | c c c c c c | c c c c c c | c }
    \multirow{3}*{Encoders} & \multicolumn{6}{c|}{NYU } & \multicolumn{6}{c|}{KITTI} & \multicolumn{1}{c}{} \\
    & RMSE$\downarrow$ & \multirow{2}*{REL$\downarrow$} & \multirow{2}*{$\delta_1\uparrow$} & TX1$\uparrow$ & Nano$\uparrow$ & MAC & RMSE$\downarrow$ & \multirow{2}*{REL$\downarrow$} & \multirow{2}*{$\delta_1\uparrow$} & TX1$\uparrow$ & Nano$\uparrow$ & MAC & Parameters \\
     & [m]  & & & [fps] & [fps] & [G] & [m] & & & [fps] & [fps] & [G] & [M] \\
     \hline
    MobileViT S & 0.549  & 0.168 & 0.763 & 13.3 & 10.5 & 1.222 & 4.673 & 0.128 & 0.825  & 5.1 & 4.1 & 3.046 & 5.9  \\
    MobilViT ReLU S & 0.521 & 0.153 & 0.790  & 13.3 & 10.5 & 1.222 & 4.789 & 0.140  & 0.815 & 5.1 & 4.1 & 3.046 & 5.9 \\
    \methodd SiLU S & 0.496 & 0.145 & 0.811  & 16.3 & 12.0 & 0.975 & 4.692 & 0.134  & 0.825 & 5.9 & 4.8 & 2.432 & 3.3 \\
    \textbf{\methodd S} & \textbf{0.471} & \textbf{0.134} & \textbf{0.831} & 16.3 & 12.0 & 0.975 & \textbf{4.603} & \textbf{0.126} & \textbf{0.829}  & 5.9 & 4.8 & 2.432 & 3.3\\
    \hline
    MobileViT XS & 0.572 & 0.171 & 0.754 & 13.5 & 13.3 & 0.815 & 4.734 & 0.133 & 0.822 & 5.9 & 5.1 & 2.032 & 2.8 \\
    MobilViT ReLU XS & 0.547 & 0.158 & 0.780  & 13.5 & 13.3 & 0.815 & 4.797 & 0.137  & 0.819 & 5.9 & 5.1 & 2.032 & 2.8 \\
    \methodd SiLU XS & 0.539 & 0.156 & 0.787 & 18.3 & 15.6 & 0.579 & 4.727 & 0.133 & 0.821  & 7.2 & 6.0 & 1.444 & 1.4 \\
    \textbf{\methodd XS} & \textbf{0.522} & \textbf{0.154} & \textbf{0.793} & 18.3 & 15.6 & 0.579 & \textbf{4.671} & \textbf{0.128} & \textbf{0.827} & 7.2 & 6.0 & 1.444 & 1.4 \\
    \hline
    MobileViT XXS & 0.615 & 0.195 & 0.715 & 17.4 & 16.9 & 0.472 & 5.211 & 0.187 & 0.761 & 14.3 & 10.7 & 1.180 & 1.8  \\
    MobilViT ReLU XXS & 0.588 & 0.176 & 0.737  & 17.4 & 16.9 & 0.472 & 5.210  & 0.171 & 0.763 & 14.3 & 10.7 & 1.180 & 1.8 \\
    \methodd SiLU XXS & 0.596 & 0.180 & 0.728 & 25.8 & 23.2 & 0.186 & 5.208 & 0.165 & 0.763  & 20.4 & 15.1 & 0.464 & 0.7 \\
    \textbf{\methodd XXS} & \textbf{0.580} & \textbf{0.174} & \textbf{0.744} & 25.8 & 23.2 & 0.186 & \textbf{5.157} & \textbf{0.156} & \textbf{0.782} & 20.4 & 15.1 & 0.464 & 0.7 \\
  \end{tabular}
\end{table*}

\begin{table*}[t]
  \centering
  \caption{Comparison between lightweight decoder architectures keeping \methodd S encoder fixed. The best scores are in bold.}
  \label{tab:decoder_comaprison}
  \begin{tabular}{ l | c c c c c c | c c c c c c | c }
    \multirow{3}*{Decoders} & \multicolumn{6}{c|}{NYU } & \multicolumn{6}{c|}{KITTI } & \multicolumn{1}{c}{} \\
    & RMSE$\downarrow$ & \multirow{2}*{REL$\downarrow$} & \multirow{2}*{$\delta_1\uparrow$} & TX1$\uparrow$ & Nano$\uparrow$ & MAC & RMSE$\downarrow$ & \multirow{2}*{REL$\downarrow$} & \multirow{2}*{$\delta_1\uparrow$} & TX1$\uparrow$ & Nano$\uparrow$ & MAC & Parameters \\
     & [m]  & & & [fps] & [fps] & [G] & [m] & & & [fps] & [fps] & [G] & [M] \\
    \hline
    NNDSConv5 \cite{fastdepth} & 0.596  & 0.174 & 0.685 & 15.4 & 11.5 & 0.869 & 5.737 & 0.164  & 0.677 & 5.5 & 4.7 & 2.166 & 3.1 \\
    NNConv5 \cite{fastdepth} & 0.562  & 0.167 & 0.761 & 14.6 & 11.3 & 1.141 & 4.895 & 0.139  & 0.818 & 5.6 & 4.5 & 2.845 & 3.6 \\
    MDSPP \cite{speed} & 0.581  & 0.169 & 0.694 & 15.1 & 11.7 & 1.004 & 5.167 & 0.157  & 0.760  & 5.7 & 4.7 & 2.503 & 3.4 \\
    \textbf{\methodd S} & \textbf{0.471} & \textbf{0.134} & \textbf{0.831} & \textbf{16.3} & \textbf{12.0} & 0.975 & \textbf{4.603} & \textbf{0.126} & \textbf{0.829}  & \textbf{5.9} & \textbf{4.8} & 2.432 & 3.3 
  \end{tabular}
\end{table*}

\subsection{Ablation study: the encoder-decoder architecture}
\label{subsec:architecture_ablation}
In this subsection we compare the performances of the encoder and the decoder components of \method; results are reported in Table~\ref{tab:encoder_comaprison} and Table~\ref{tab:decoder_comaprison}, respectively.
In particular, the first analysis highlights the contribution of the novel \methodd block for each configuration (S, XS, and XXS) while keeping \methodd decoder fixed. 
The second analysis focuses on the use of alternative decoders with respect to the default \methodd decoder, such as NNDSConv5, NNConv5~\cite{fastdepth} and MDSPP~\cite{speed} using \methodd S encoder since it is the encoder that shows the best performances in the evaluated metrics. 


\vspace{1em}
\textbf{Encoder} architectures are compared in Table~\ref{tab:encoder_comaprison}, reporting a one-to-one comparison between \methodd encoder and the MobileViT; evaluating the effects of two different activation functions (ReLU, SiLU).
From the obtained results, we highlight that \methodd encoder (in bold) achieves better depth estimation in all the proposed variants, as well as when compared with the same activation function, using fewer trainable parameters and a reduced number of MAC operations.
In particular, when compared with the MobileViT, \methodd achieves an average improvement of $10\%$, $14\%$ and $6\%$ on RMSE, REL, and $\delta_1$ metrics in the indoor dataset and of $2\%$, $7\%$ and $2\%$ respectively on the outdoor dataset.
Based on those findings, the overall estimation contribution of the proposed encoder over the three configurations is equivalent to $7\%$, which almost $3\%$ is due to the use of ReLU activation function with respect to SiLU.
Moreover, regarding MAC operations we obtain a reduction of  $20\%$, $29\%$, and $60\%$ with respect to the corresponding MobileViT variants (S, XS, XXS), while the fps improvements are respectively $16\%$ fps, $22\%$ fps, and $32\%$ on the NVIDIA Jetson TX1 and of $16\%$ fps, $15\%$ fps, and $28\%$ fps over the NVIDIA Jetson Nano.


In light of the previous experiments, we can state that all \methodd variants show good accuracy and frequency performances on the NYU Depth v2, while in the case of KITTI dataset \methodd XXS variant should be preferred in order to get a reasonable inference speed.
Focusing on the timings, the \methodd XXS variant shows the fastest inference speed, with reasonable results also on high resolution images of KITTI dataset, avoiding the needing of cropping or downscaling the original images.


\vspace{1em}
\textbf{Decoders} architectures are reported in Table~\ref{tab:decoder_comaprison}, comparing \methodd decoder and those of other lightweight models; we used the \methodd S encoder as baseline.
\methodd decoder achieves an RMSE improvement of $16\%$ and $19\%$ on NYU Depth v2 dataset and of $6\%$ and $11\%$ on KITTI dataset with respect to NNConv5 and MDSPP models.
Furthermore, we compare \methodd decoder with the NNDSConv5 \cite{fastdepth}, a variant of the NNConv5 that takes advantage of depthwise separable convolution to reduce the computational cost. 
Our encoder-decoder architecture is able to achieve higher speed and a significant improvement in all the estimation metrics with comparable MAC operations with respect to NNDSConv5.
Finally, when compared with the NNConv5 decoder, ranked second in our analysis, the proposed structure is able to achieve an overall improvement equal to $12\%$ over the two scenarios.
Moreover, it can be noticed that the decoder has little influence on the inference frequency; however, \methodd decoder still shows the best fps on the two hardwares (e.g. $11\%$ of \methodd S compared to NNConv5 on the TX1 hardware and NYU Depth v2 dataset).
The overall MAC operations decrement with respect to NNConv5 and MDSPP decoders
is equal to $15\%$ on the same configuration as before, suggesting that the optimized \methodd decoder is able to produce more accurate estimations while using less operations.


\begin{figure*}[t]
    \centering
    \includegraphics[width=\linewidth]{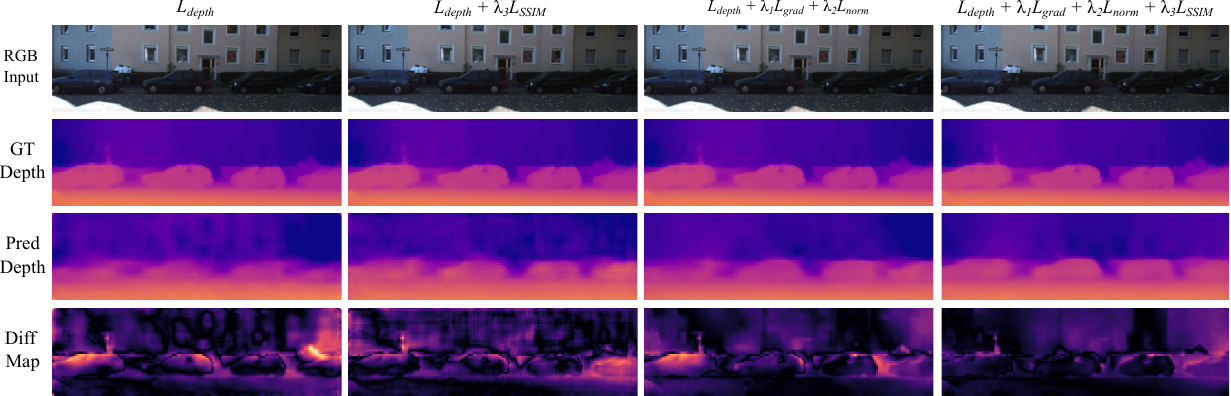}
    \caption{Qualitative comparison of a predicted frame taking into account different loss components. 
    For a better visualization, we apply to the depth images and to the difference maps uniform colormaps with the same depth range. Precisely, in the ground truth (GT) and predicted depth maps (Pred) a lower color intensity corresponds to further distances, while in the difference map (Diff $= |$GT $-$ Pred$|$) a lower color intensity corresponds to a smaller error.}
    \label{fig:loss_qualitative_comparison}
\end{figure*}

\begin{table*}[!t]
  \centering
  \caption{The effect of each \textit{balanced loss function} components on the \methodd S over the considered metrics. The best scores are in bold.}
  \label{tab:loss_comparison}
  \begin{tabular}{ l | c c c | c c c }
    \multirow{3}*{Loss Components} & \multicolumn{3}{c|}{NYU } & \multicolumn{3}{c}{KITTI }\\
    & RMSE$\downarrow$ & \multirow{2}*{REL$\downarrow$} & \multirow{2}*{$\delta_1\uparrow$} & RMSE$\downarrow$ & \multirow{2}*{REL$\downarrow$} & \multirow{2}*{$\delta_1\uparrow$} \\
     & [m]  & & & [m]  & & \\
     \hline
    $L_{depth}$ & 0.582 & 0.185 & 0.736 & 5.637  & 0.183 & 0.741\\
    $L_{depth}$ + $\lambda_3 L_{SSIM}$ & 0.544 & 0.161 & 0.774 & 5.526  & 0.198 & 0.731 \\
    $L_{depth}$ + $\lambda_1 L_{grad}$ + $\lambda_2 L_{norm}$ & 0.522  & 0.153 & 0.792 &  5.285 & 0.166 & 0.744 \\
    $L_{depth}$ + $\lambda_1 L_{grad}$ + $\lambda_2 L_{norm}$ + $\lambda_3 L_{SSIM}$ \textbf{(BLF)} & \textbf{0.471} & \textbf{0.134} & \textbf{0.831}  &  \textbf{4.603} & \textbf{0.126} & \textbf{0.829} \\
  \end{tabular}
\end{table*}

\subsection{Ablation study: loss function}
\label{subsec:loss_ablation}
\begin{figure}[h]
    \centering
    \includegraphics[width=\linewidth]{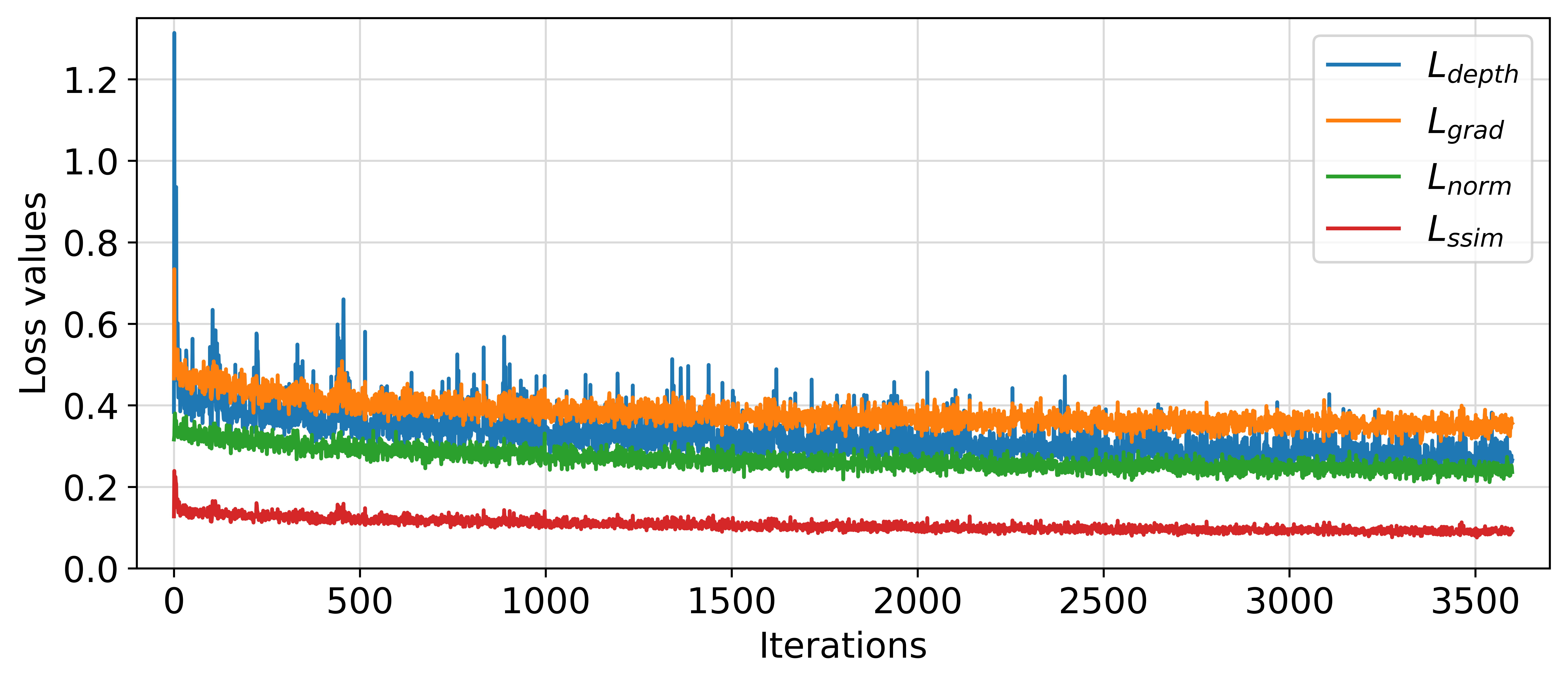}
    \caption{Plot of the individual loss components  composing the \textit{balanced loss function} in the first ten epochs, i.e. almost 3600 iterations.}
    \label{fig:loss_trends}
\end{figure}

In this subsection we analyze the impact of the different components of the proposed \textit{balanced loss function} introduced in Section~\ref{subsec:loss_function}. 
\methodd S architecture is used as a baseline model.
The quantitative and qualitatively comparisons are provided in Table~\ref{tab:loss_comparison} and Figure~\ref{fig:loss_qualitative_comparison} respectively, while Figure~\ref{fig:loss_trends} shows the converging trends of each introduced component, referring to $L_{depth}$ (blue), $L_{grad}$ (orange), $L_{norm}$ (green) and $L_{SSIM}$ (red).

The curves shape show that the initial loss contribution is mostly attributed to the $L_{depth}$ and $L_{grad}$, while the contributions of the $L_{SSIM}$ and $L_{norm}$ penalize from start to finish structural and high-level details prediction errors.

The $L_{depth}$ component showed to be fundamental for the training convergence, thus it is applied on every experiment of Table~\ref{tab:loss_comparison}. 
The obtained results demonstrate that each loss component is crucial to get the final \methodd performance, balancing the reconstruction of the entire image and of edges details. In fact, the loss formulation in the second row focuses only on the overall image, failing at reaching satisfying results. At the same time, the third row shows a typical loss exploited in~\cite{acc_obj_loss}  focusing on edge details but not taking into account the image structure similarity, thus producing an unbalanced loss achieving a worse result with respect to the proposed one, which is able to obtain the lowest estimation error by balancing all the components.
In detail, the BLF achieves an improvement of $10\%$, $12\%$, and $5\%$ for RMSE, REL and $\delta_1$ metrics on NYU dataset, and a boost of $13\%$, $24\%$, and $10\%$ over the KITTI dataset compared to~\cite{acc_obj_loss}.

Moreover, to better show the qualitative contribution of each loss component, provided in Figure~\ref{fig:loss_qualitative_comparison} the estimated depth under the four analyzed configurations given an input sample from KITTI dataset.
Based on such example, we can observe a similar behaviour to the one found in Figure~\ref{fig:loss_trends} and Table~\ref{tab:loss_comparison}: the $L_{depth}$ component is fundamental for a correct image reconstruction while the weighted addition of specific loss components ($\lambda_1 L_{grad}$, $\lambda_2 L_{norm}$, $\lambda_3 L_{SSIM}$) can quantitatively and qualitatively improve the final estimation.
This improvement may also be noticed by observing the predicted frames from left to right, where the object details and the overall estimation increase significantly as difference maps darken.

Therefore, we can conclude that the proposed \textit{balanced loss function} can successfully enhance the training process, while each component can effectively contribute to more accurate estimations, hence enhancing the entire framework. 
Precisely, the overall quantitative contribution of the \textit{balanced loss function} over the two scenarios is equal to $25\%$ when compared with $L_{depth}$, and $13\%$ with respect to the loss formulation used in~\cite{acc_obj_loss}.

\subsection{Ablation study: data augmentation}
\label{subsec:DA_ablation}
In this ablation study, we evaluate the performances of the proposed data augmentation strategy in comparison with standard MDE data augmentation. 
We report in Table~\ref{tab:aug_comparison} the quantitative results of \textit{shifting strategy} (\textit{C} shift, \textit{D} shift) and the \textit{default } DA (flip, random crop and channel swap) and the combinations of the two.
The proposed \textit{shifting strategy} (last row) achieves, on \methodd S architecture, an improvement of $8\%$, $6\%$, and $2\%$ over the RMSE, REL and $\delta_1$ on the NYU Depth v2 dataset, and of $6\%$, $2\%$, and $1\%$ over the KITTI dataset. 
On the other hand, the single use of the \textit{C} shift or \textit{D} shift with the \textit{default} augmentation does not lead to an improvement in the final estimation, resulting in equivalent or slightly worst final prediction.
Then, the overall improvement of the \textit{shifting strategy} over the two scenarios is equal to $4\%$ with respect to the \textit{default} data augmentation policy.

\begin{table*}[t]
  \centering
  \caption{Comparison between different augmentation strategies. The \textit{default} policy comprises the flip, random crop and channel swap while the others represent the different components of the \textit{shifting strategy} described in Section \ref{subsec:combined_augmentation}. The reference model is \methodd S. The best scores are in bold.}
  \label{tab:aug_comparison}
  \begin{tabular}{ l | c c c | c c c }
    \multirow{3}*{Augmentation Components} & \multicolumn{3}{c|}{NYU } & \multicolumn{3}{c}{KITTI }\\
    & RMSE$\downarrow$ & \multirow{2}*{REL$\downarrow$} & \multirow{2}*{$\delta_1\uparrow$} & RMSE$\downarrow$ & \multirow{2}*{REL$\downarrow$} & \multirow{2}*{$\delta_1\uparrow$} \\
     & [m]  & & & [m]  & & \\
     \hline
    \textit{default} & 0.511 & 0.143 & 0.813 & 4.839 & 0.128 & 0.826 \\
    \textit{default} + \textit{C} shift & 0.506 & 0.143 & 0.815 & 4.897 & 0.136 & 0.810 \\
    \textit{default} + \textit{D} shift & 0.585 & 0.144 & 0.805  & 4.938 & 0.141 & 0.804 \\
    \textit{default} + \textit{C} shift + \textit{D} shift \textbf{(our)} & \textbf{0.471} & \textbf{0.134} & \textbf{0.831}  & \textbf{4.603} & \textbf{0.126} & \textbf{0.829} \\
  \end{tabular}
\end{table*}

\begin{figure}[t]
    \centering
    \includegraphics[width=\linewidth]{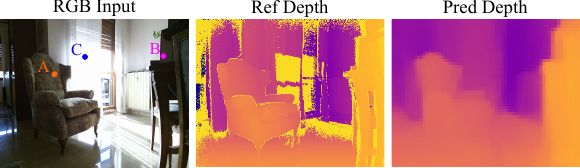}
    \caption{\methodd application in a real-case scenario. Missing depth measurements of reference (Ref) depth are shown as yellow pixels. 
    A uniform colormap, with the same depth range, has been applied to the depth maps.
    Points A (armchair), B (box), and C (curtain) on the RGB frame indicates object used for quantitative comparison.}
    \label{fig:room_test}
\end{figure}

\subsection{Real-case scenario}
\label{subsec:real_case_usage}
One of the main objectives of exploring lightweight deep learning solutions is to close the gap between computer vision and practical applications, where the proposed models may be integrated as perception systems, such as robotic systems, thus taking into account possible hardware limitations. 
Therefore, in this subsection, we present an example of a real-case application in which METER is used to estimate the depth scene obtained from a generic camera image. 
We used a Kinetic V2 to measure the reference depth of the scene.
The extracted acquisition is reported in Figure~\ref{fig:room_test}.

Qualitatively comparing the reference depth and the estimated one, we can notice a less sharp prediction, which can be mainly attributed to the lower working resolution that ensures high frame rates on edge devices. 
However, the object shapes are still adequately defined, and the overall estimation is visually comparable with the reference frame.

Moreover, in order to perform a quantitative analysis, we compute the average error of three salient objects that appear in the input frame (RGB Input), which are point A for the armchair, point B for the box and point C for the curtain.
The estimation error for the first two points (A and B) is almost equal to $0.79$m, respectively. 
The obtained value is related to the fact that we are working in a challenging open-set scenario with different statistics with respect to the training set.
On the other hand, by analyzing point C, we can identify one of the main drawbacks of active depth sensing, i.e. missing or incorrect depth measurements under particular lighting conditions. 
In this scenario, although the estimated depth error is unknown, most likely due to the intense light source directed towards the camera sensor, our model can still correctly identify and estimate the area as a single surface.

\section{Conclusions}
\label{sec:conclusion}
In this work, we propose \method, a MDE architecture characterized by a novel lightweight vision transformer model, a multi-component loss function and a specific data augmentation policy. 
Our method exploits a lightweight encoder-decoder architecture characterized by a transformer \methodd block, which is able to improve the final depth estimation with a small number of computed operations, and a fast upsampling block employed in the decoder.
\methodd achieves high inference speed over low-resource embedded hardwares such as the NVIDIA Jetson TX1 and the NVIDIA Jetson Nano. 
Moreover, \methodd architecture in its three configurations is able to outperform previous state of the art lightweight related works. 
Thanks to the obtained performances on inference frequency and accuracy in the estimation, such proposed architectures can be good candidate to work on  multiple MDE scenarios and real-world embedded applications.
Precisely, \methodd S outperforms the accuracy of state of the art lightweight methods over the two datasets,  \methodd XS represents the best trade-off between inference speed and estimation error, and \methodd XXS reaches a high inference frequency, up to $25.8$ fps, on the two hardwares at the cost of a small increment in the estimation error.

The obtained results and the limited MAC operations of the proposed network demonstrate that our framework could be valuable in a variety of resource-constrained applications, such as autonomous systems, drones, and IoT.
Moreover, we also test \methodd in a real-case scenario with a frame captured by a generic camera achieving a reasonable estimation error.

Finally, \methodd architecture could be a valuable starting point for future studies, in order to get real-time inference frequency on high resolution images, as well as building transformer architectures to take advantage of the attention mechanism both in encoder and decoder structures.

\section*{Acknowledgments}
This work was partially supported by the Sapienza University of Rome project 2022-2024 ``A Novel Vision-based detection system for the control of the ectoparasitic mite Varroa destructor in honey bee colonies''.

\bibliographystyle{IEEEtran}
\bibliography{main}

\newpage
\begin{IEEEbiography}[{\includegraphics[width=1in,height=1.25in,clip,keepaspectratio]{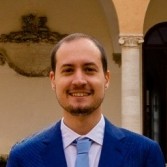}}]{LORENZO PAPA} is a Ph.D. student in Computer Science Engineering. He collaborates with the AlcorLab in the DIAG department, University of Rome Sapienza, Italy. He received the B.S. degree in Computer and Automation Engineering and the M.S. degree in Artificial Intelligence and Robotics from the University of Rome La Sapienza, Italy, in 2019 and 2021, respectively.
His main research interests are Deep Learning, Computer Vision and Cyber Security.
\end{IEEEbiography}

\begin{IEEEbiography}[{\includegraphics[width=1in,height=1.25in,clip,keepaspectratio]{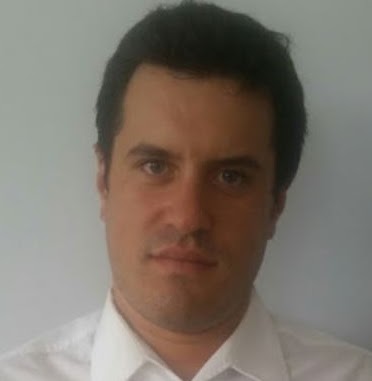}}]{PAOLO RUSSO} is an Assistant Researcher at AlcorLab in DIAG department, University of Rome Sapienza, Italy. 
He received the B.S. degree in Telecommunication Engineering from Università degli studi di Cassino, Italy, in 2008, and the M.S. degree in Artificial Intelligence and Robotics from University of Rome La Sapienza, Italy, in 2016. 
He received Ph.D. degree in Computer Science from University of Rome La Sapienza in 2020.
From 2018 to 2019, he has been a researcher at Italian Institute of Technology (IIT) in Tourin, Italy.
His main research interests are Deep Learning, Computer Vision, Generative Adversarial Networks, Reinforcement Learning.
\end{IEEEbiography}

\begin{IEEEbiography}[{\includegraphics[width=1in,height=1.25in,clip,keepaspectratio]{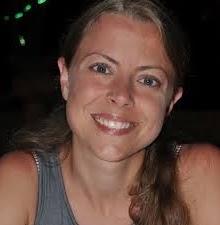}}]{IRENE AMERINI} (M’17) received the Laurea degree in computer engineering and the Ph.D. degree in computer engineering, multimedia, and telecommunication from the University of Florence, Italy, in 2006 and 2010, respectively. 
She is currently an Assistant Professor with the Department of Computer, Control, and Management Engineering A. Ruberti, Sapienza Univeristy of Rome, Italy. She was a Visiting Scholar with Binghamton University, NY, USA, in 2010, and a Visiting Research Fellow of Charles Sturt University, Australia, in 2018, with a fellowship offered by the Australian Government Department of Education and Training, through the Endeavour Scholarship \& Fellowship Program. 
Her main research activities include digital image processing, multimedia content security technologies, secure media, and multimedia forensics. She is a member of the IEEE Information Forensics and Security Technical Committee and the EURASIP TAC Biometrics, Data Forensics, and Security and the IAPR TC6 - Computational Forensics Committee. 
She has received the Italian Habilitation for an Associate Professor in telecommunications and computer science. She is a Guest Editor of several international journals. She is an Associate Editor of IEEE ACCESS, Journal of Electronic Imaging and Journal of Information Security and Applications.
\end{IEEEbiography}

\vfill

\end{document}